\documentclass[runningheads]{llncs}
\usepackage{graphicx}
\usepackage{booktabs}
\usepackage{xcolor}
\usepackage{multirow}
\usepackage{arydshln}
\usepackage{paralist}
\usepackage{hyperref}

\newcommand{\data}{\textsc{FloCo}}
\newcommand{\task}{\textsc{Flow2Code}}
\newcommand{\model}{\textsc{FloCo-T5}}

\begin{document}

\title{Towards Making Flowchart Images\\ Machine Interpretable}

\author{Shreya Shukla\orcidID{0009-0002-9029-2547} \and
Prajwal Gatti\orcidID{0000-0002-6554-3132} \and
Yogesh Kumar\orcidID{0009-0009-4363-5317} \and
Vikash Yadav\orcidID{0009-0000-3245-8137} \and
Anand Mishra\orcidID{0000-0002-7806-2557}}

\authorrunning{Shukla et al.}

\institute{
Vision, Language and Learning Group (VL2G)\\
IIT Jodhpur, India\\
\email{\{shukla.12, pgatti, kumar.204, yadav.41, mishra\}@iitj.ac.in}}

\maketitle
\begin{abstract}  
Computer programming textbooks and software documentations often contain flowcharts to illustrate the flow of an algorithm or procedure. Modern OCR engines often tag these flowcharts as graphics and ignore them in further processing. In this paper, we work towards making flowchart images machine-interpretable by converting them to executable Python codes. To this end, inspired by the recent success in natural language to code generation literature, we present a novel transformer-based framework, namely \model{}. Our model is well-suited for this task, as it can effectively learn semantics, structure, and patterns of programming languages, which it leverages to generate syntactically correct code. We also used a task-specific pre-training objective to pre-train \model{} using a large number of logic-preserving augmented code samples. Further, to perform a rigorous study of this problem, we introduce the \data{} dataset that contains 11,884 flowchart images and their corresponding Python codes. Our experiments show promising results, and \model{} clearly outperforms related competitive baselines on code generation metrics. We make our dataset and implementation publicly available\footnote{\url{https://vl2g.github.io/projects/floco}}.

\end{abstract}

\section{Introduction}
\begin{figure}[t!]
\centering
\includegraphics[width=\textwidth]{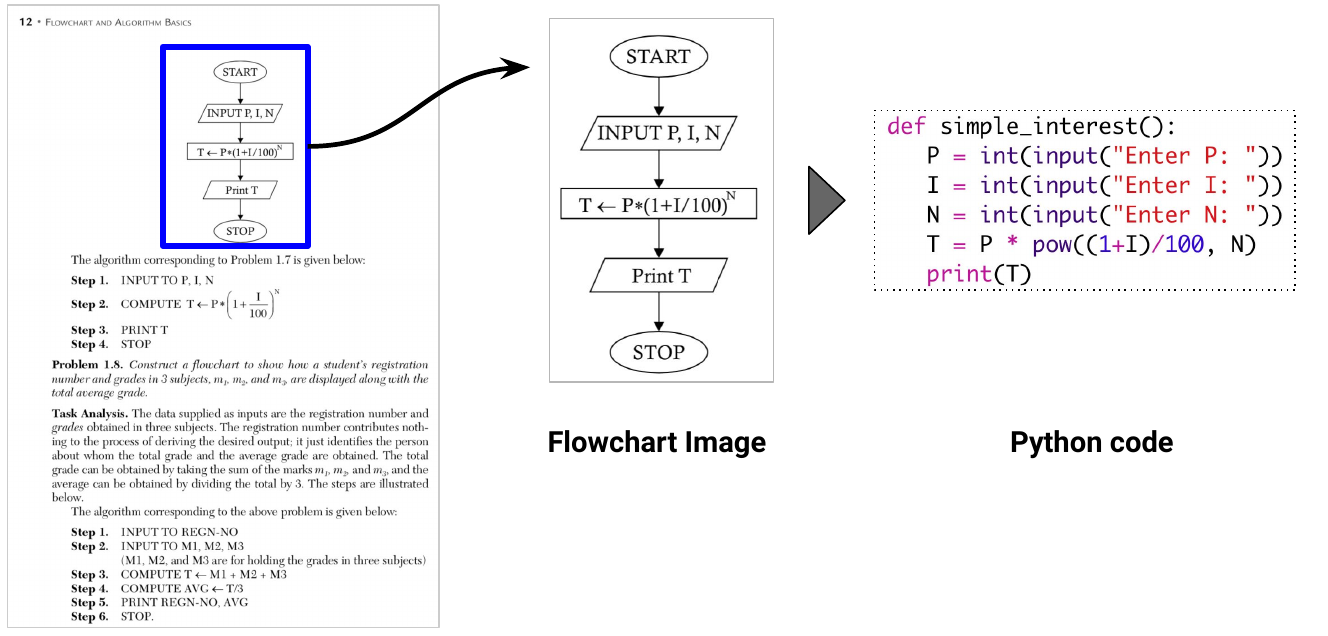}
\caption{\textbf{\task{}}. A scanned document from a programming text book~\cite{chaudhuri2020flowchart} containing a flowchart is shown here. Our aim is to convert flowchart images to executable computer programs. We scope ourselves to cropped flowchart images and Python codes in this work. \label{fig:text_charts}}
\end{figure}
Flowcharts are widely used across documents to represent algorithms, processes, or workflows in a graphical manner and provide a clear and concise understanding of complex processes. They contain short textual commands or conditions inside various intent-specific shapes, e.g., diamond for decision-making block, rhomboid for input, and output. These shapes are connected with directed or undirected arrows to define a sequential flow of information and processing. In computer programming textbooks and software documentation, flowcharts are more often used as a program-planning tool to communicate the complex logic of programs and keep track of the data flow through a process, as shown in figure~\ref{fig:text_charts}. These visual depictions help beginners in programming to focus on formulating the logic behind the program while ignoring the intricacies of the syntax of different programming languages. Machine interpretation of these flowcharts followed by automatic code generation would not only help school students and people from non-software engineering backgrounds but also speed up software development. In order to make these flowchart images machine-interpretable, we study the problem of automatically converting flowchart images to a computer program (code) in a high-level language. This problem is referred to as \task{}~\cite{herrera2017flow2code}. Despite its practical importance and utility, \task{} has not been rigorously explored in the literature. 

There is no existing large-scale dataset in flowchart-to-code literature for performing a rigorous experimental evaluation of \task{} task. To fill this gap, we introduce the first dataset, namely \data{}. The \data{} contains 11,884 flowchart images along with corresponding Python codes. Inspired by the success of transformer-based approaches in natural language and code generation tasks~\cite{ahmad2021unified,akermi2020transformer,brown2020language,kenton2019bert,raffel2020exploring}, we present \model{} -- a novel framework to convert flowchart images to the Python code. In our proposed architecture, we first convert the flowchart images into a sequence encoding by automatically detecting different shapes and reading text using off-the-shelf OCR engines~\cite{easy_ocr,li2021trocr}. Then, to adapt our transformer model to this novel domain, we pre-train it on the masked token modeling objective using a large number of logic-preserving data-augmented code samples. This pre-training step helps the model to understand the structure and semantics of the programming language as well as the flowchart encoding. Finally, we fine-tune the model on train split as a sequence-to-sequence generation problem where flowchart encoding and expected Python code are used as input and output sequence, respectively. We conducted extensive experiments with the sequence encoding of the flowchart images (as shown in Table~\ref{table:encodings}) and compared the code generation performance of our model against competitive baselines, namely Vanilla Transformer~\cite{vaswani2017attention}, BART~\cite{lewis2020bart}, PLBART~\cite{ahmad2021unified} and CodeT5~\cite{wang2021codet5}. Our experiments show that \model{} outperforms all other baselines on different code generation metrics, showing the efficacy of the proposed pre-training objective and data augmentation techniques. Qualitative results and further analysis (Figures ~\ref{fig:qaul_res} and ~\ref{fig:handwritten_res}) demonstrate that our model effectively learns the structure and pattern of programming languages and the logical data flow and generates syntactically correct code corresponding to the flowchart images.

\noindent\textbf{Contributions:} The major contributions of this work are three folds:

\begin{enumerate}
    \item We study the \task{} task in a ``large-scale setting" and introduce an accompanying dataset -- \data{} containing 11,884 flowchart images and corresponding Python codes. This dataset shall enable future research in this under-explored space (Section~\ref{dataset}).
    
    \item We propose a novel framework viz. \model{} to address the task in hand, which involves generating flowchart encodings, pre-training CodeT5 on the task-specific objective with augmented codes, and finally fine-tuning for the code generation task (Section~\ref{method}).
       
    \item We conducted extensive experiments with various baselines and proposed task-specific code augmentation and pre-training strategy. We achieve BLEU, CodeBLEU, and exact match scores of $67.4$, $75.7$, and $20.0$, respectively. Towards the end, we show that our model can be adopted to hand-drawn flowchart images as well (Section~\ref{experiment}).
    \end{enumerate}

\section{Related Work}
\subsection{Flowchart Understanding}
There have been several attempts to build software for flowchart-to-code conversion, such as authors in~\cite{cook2014flowgorithm}, and ~\cite{Supaartagorn2017webapp} introduced interactive user interfaces to convert flowcharts to codes on-the-fly in various programming languages. These rule-based approaches, however, impose restrictions and do not support the conversion for offline flowchart images like ours. 
In~\cite{wu2011research}, a platform was designed to recognize flowcharts and convert them to ANSI-C code using structure identification. In~\cite{herrera2017flow2code}, a method was proposed for handwritten flowcharts, using rule-based techniques for preprocessing and generating pseudo code. In~\cite{Carton2013flow}, improved results were achieved in flowchart recognition by combining statistical and structural information. In~\cite{Schäfer2019ArrowRCNN}, the Faster RCNN object detection system was extended with an arrow keypoint predictor to recognize handwritten flowcharts. In~\cite{Fang2022DrawnNet}, DrawnNet was proposed, a keypoint-based detector for handwritten diagram recognition.

A recent work~\cite{tannertflowchartqa} introduced a novel benchmark and dataset for question-answering over flowcharts. However, their flowchart images are unsuited for programming tasks and can not be used for our problem. The work closest to our setting is ~\cite{betancourt2022handwritingflowchartCNN}, which targets the digitization of handwritten flowchart images with Faster RCNN and OCR-techniques, followed by converting them to codes in C programming language using a CNN-LSTM based model. In this task, the authors propose a dataset of 775 handwritten flowchart images in Spanish and English languages. However, this dataset is unsuited for \task{} as it only consisted of hand-drawn flowchart images, with many samples consisting of only box drawings with no text, and the corresponding C codes were publicly unavailable. In this work, we consider \task{} as a sequence-to-sequence generation problem and address it using a state-of-the-art transformer-based technique in a data-driven manner. Further, we curate a dataset of 11.8K samples containing both digitized and handwritten flowchart images along with their corresponding Python codes to provide a more suitable benchmark for this task.

\subsection{Large-scale pre-trained Language Models}
The introduction of the transformer~\cite{vaswani2017attention} architecture has brought a remarkable revolution in natural language processing. Further, to deal with the scarcity of labeled data and build a general-purpose model for a wide range of NLP applications, Radford et al.~\cite{radford2018improving} proposed GPT, which is based on a transformer-decoder and pre-trained with an unlabeled pool of data in a self-supervised fashion. However, it follows a unidirectional autoregressive approach and is not suitable for tasks utilizing information from the entire sequence. Kenton et al. introduced BERT~\cite{kenton2019bert}, a transformer-encoder-based method trained in a similar self-supervised fashion. BERT~\cite{kenton2019bert} follows a bidirectional autoencoder nature and is unsuitable for generation tasks that utilize information from the previously generated tokens in the sequence. To deal with the shortcomings of GPT~\cite{radford2018improving} and BERT~\cite{kenton2019bert}, Lewis et al. introduced BART~\cite{lewis2020bart}, a denoising autoencoder that uses a bidirectional encoder and an auto-regressive decoder. These large-scale language models are often fine-tuned with a small set of labeled data for the supervised downstream task. In general, there are other well-explored pre-trained transformer-based methods such as T5~\cite{raffel2020exploring}, MASS~\cite{song2019mass}, ELECTRA~\cite{clark2020electra}, and RoBERTa~\cite{liu2019roberta}. In this work, we utilize CodeT5~\cite{wang2021codet5}, which adopts the encoder-decoder-based transformer model viz. T5~\cite{raffel2020exploring}, and is pre-trained on programming language data.  
 
\subsection{Language Modeling for Code Generation}
 A significant amount of effort has been invested in automating software engineering using deep learning. Recent work has focused on transferable representations rather than task-specific ones. Pre-trained NLP models like BERT~\cite{kenton2019bert}, GPT~\cite{radford2018improving}, and BART~\cite{lewis2020bart}  have demonstrated transferability to programming languages, yielding positive results for a range of code-related tasks. 
 
Feng et al.~\cite{feng2020codebert} introduced CodeBERT, which utilized BERT architecture pre-trained on programming language and natural language used in the software development domain, with masked language modeling objective. Guo et al.~\cite{guo2020graphcodebert} proposed GraphCodeBERT as an improvement upon CodeBert by leveraging dataflow in source code through two additional pre-training tasks, predicting code structure edges, and aligning representations between source code and code structure. Ahmad et al.~\cite{ahmad2021unified} introduced PLBART, a bidirectional and autoregressive transformer pre-trained on unlabeled natural language and programming language data, with denoising autoencoding objective, where the noising strategies employed were token masking, token deletion, and text infilling. Wang et al. ~\cite{wang2021codet5} proposed CodeT5 by extending T5~\cite{wang2021codet5} to programming languages. Similar to PLBART~\cite{ahmad2021unified}, it is a unified encoder-decoder transformer model, but it has task-specific fine-grain pre-training objectives such as masked span prediction, identifier tagging, masked identifier prediction, and bimodal dual generation objectives. As CodeT5~\cite{wang2021codet5} has the advantage of task-specific pre-training strategies, we adopted it for our main method. We generated sequential encodings from flowchart images to treat \task{} as a sequence-to-sequence problem. We pre-trained the CodeT5 model with masked token modeling objective on a large number of logic-preserved augmented codes. Finally, we fine-tuned the pre-trained model for code generation. 

\begin{table}[t!]
    \centering
    \scriptsize
      \caption{Statistics of the \data{} dataset. \label{table:data_stats}}
      \label{table:main-results}
      \begin{tabular}{l r}
        \toprule
        \textbf{Property}~~~~~&~~~~~\textbf{Value} \\
        \midrule
        Total number of samples & 11,884 \\
        Avg. length of the program (in tokens) & 46 \\
        Avg. length of the program (in lines) & 4.6 \\
        \midrule
        Train set size & 10,102 \\
        Test set size & 1,188 \\
        Validation set size & 594 \\
        \bottomrule
      \end{tabular}
\end{table}

\begin{figure}[t!]
\includegraphics[width=\textwidth]{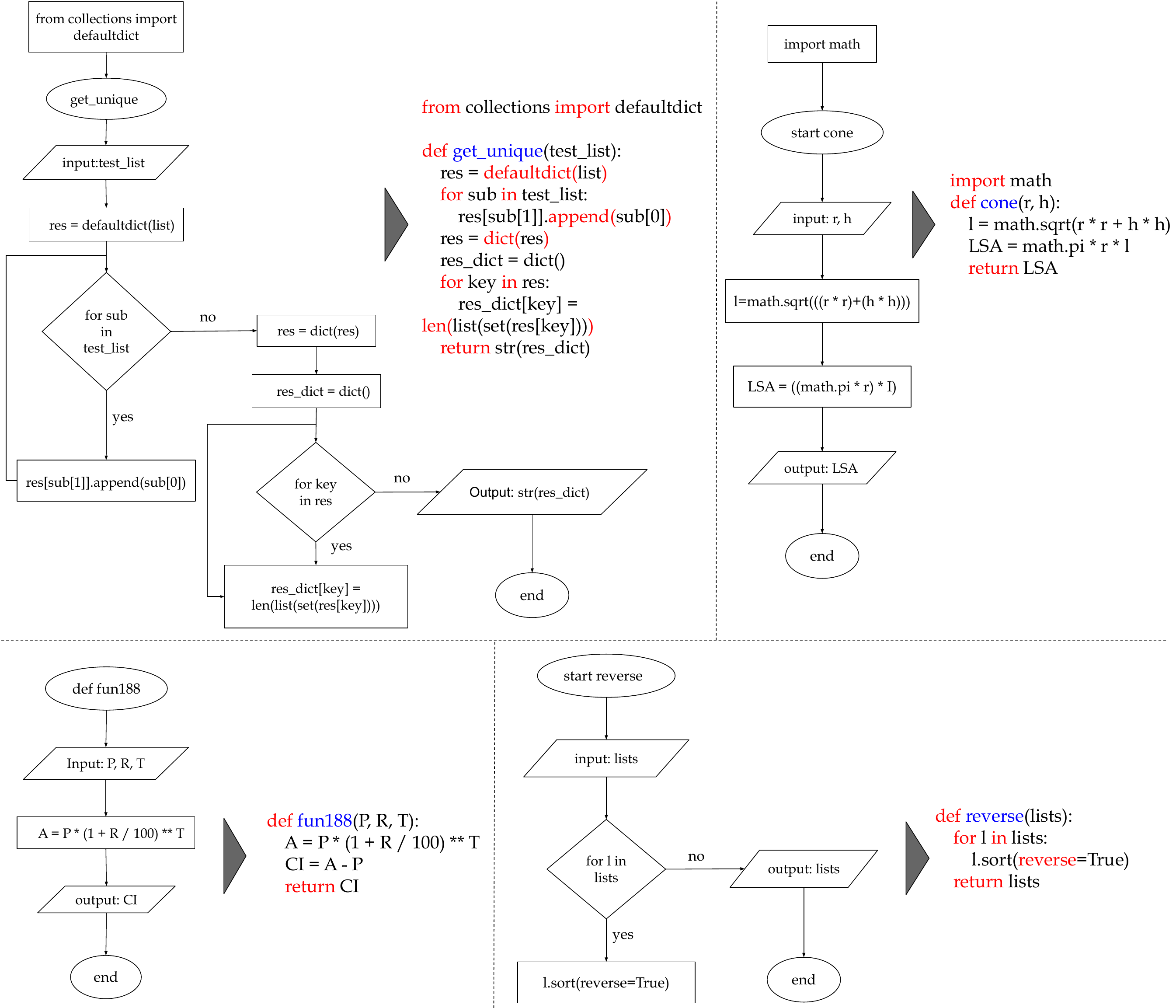}
\caption{Samples from the proposed \data{} dataset. Each flowchart image is associated with the corresponding ground truth code.}  \label{fig:datset_eg}
\end{figure}

\section{\data{}: A novel dataset for \underline{Flo}wchart image to python \underline{Co}de conversion}
\label{dataset}
We introduce a novel large-scale dataset for \underline{Fl}owchart images to python \underline{Co}de conversion. We refer to this dataset as \data{}. It contains 11,884 flowchart images and corresponding python codes. A selection of representative examples from the \data{} dataset is depicted in Figure~\ref{fig:datset_eg}.  We make \data{} publicly available for download~\footnote{\url{https://vl2g.github.io/projects/floco/}}. 

Flowchart-related research has been under-explored in the literature. However, there exist some related datasets such as (a) OHFCD dataset~\cite{OHFCD} has 419 handwritten flowcharts; however, it does not contain the corresponding codes as their focus is reading handwritten flowchart images and not code generation, (b) a more recent dataset namely FlowchartQA~\cite{tannertflowchartqa} introduces a synthetic dataset for question answering and reasoning on flowcharts. (c) in ~\cite{betancourt2022handwritingflowchartCNN}, authors introduced a collection of 775 handwritten flowchart images and corresponding C programming languages. However, codes for this dataset are not publicly available. Our new dataset, viz. \data{} has been introduced in this work to fill the research gap in the literature.

The \data{} dataset contains 11,884 flowchart images and python code pairs. The dataset has been generated by writing a few codes from scratch and gathering and cleaning codes from the MBPP (Mostly Basic Python Programs)~\cite{austin2021program}, and code-to-text dataset of CodeXGleu~\cite{lu2021codexglue} datasets. The digitized flowchart images corresponding to the codes are generated using the pyflowchart\footnote{\url{https://pypi.org/project/pyflowchart/}} and diagrams\footnote{\url{https://diagrams.mingrammer.com/}} libraries. \data{} is divided into train, test, and validation sets following an 85:10:5 ratio split respectively. Our data comprises a diverse collection of Python programs spanning a spectrum of complexity and uniqueness in their designated tasks. A few examples of these designated tasks include \textit{calculating the $N^{th}$ Fibonacci number, determining binomial coefficients, checking if a binary tree is balanced}, and \textit{finding $n^{th}$ Catalan number}. Detailed tatistics related to \data{} are provided in Table~\ref{table:data_stats}.

\begin{figure}[t!]
\includegraphics[width=\textwidth]{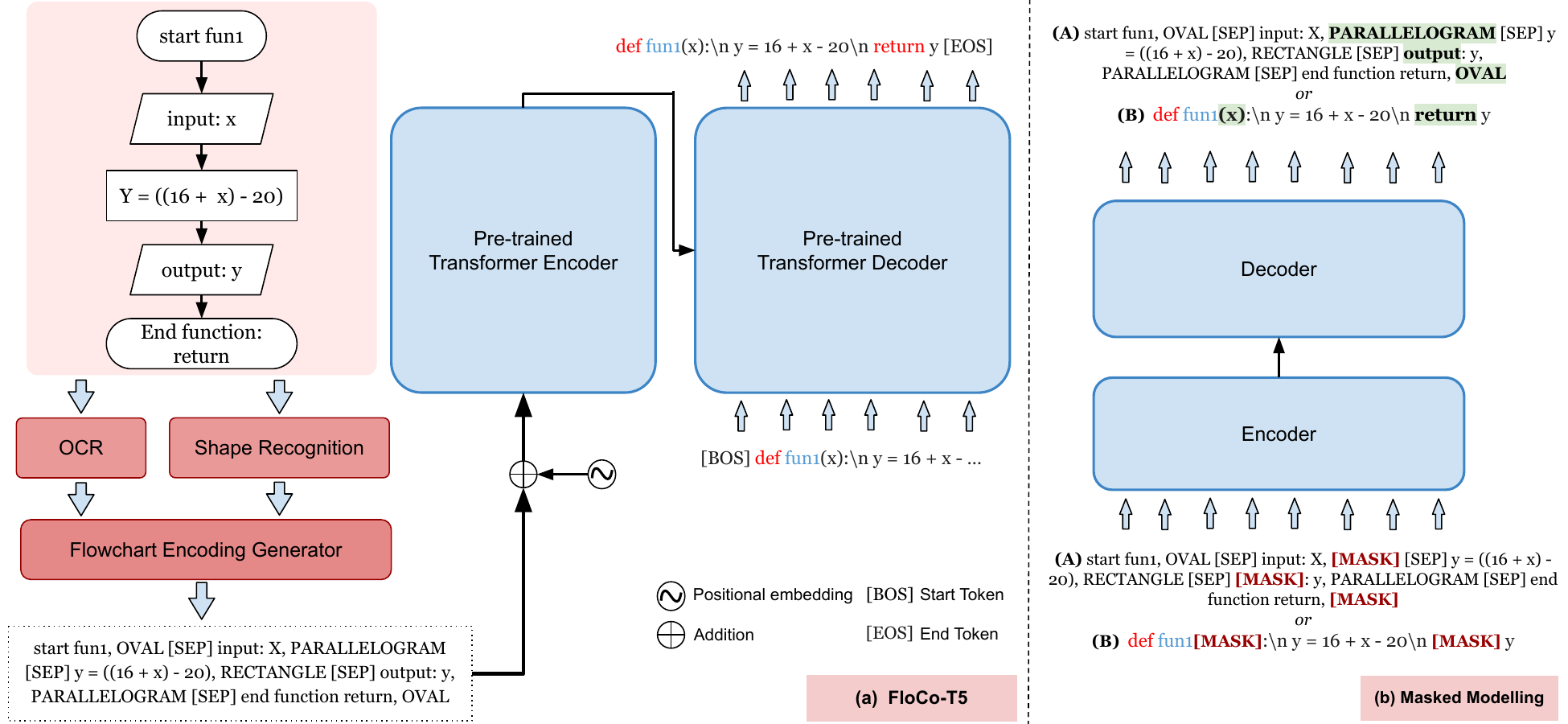}
\caption{\textbf{Overview of the proposed method, viz. \model{}:} \textbf{(a).} Given flowchart image is converted into sequence encoding using the off-the-shelf OCR techniques. The encoder of the pre-trained CodeT5 model takes flowchart encoding added with positional encodings as input. The Decoder initially takes the start token as input and has access to encoder output,  then autoregressively generates the expected code word by word. \textbf{(b).} Shows the token mask modeling. Before fine-tuning, CodeT5 is pre-trained on a token mask modeling task, where some tokens of flowchart encoding are masked and reconstructed by the decoder in an unsupervised learning fashion \textbf{[Best viewed in color].}} \label{fig:model}
\end{figure}

\section{Proposed Approach}
\label{method}
The goal of \task{} is to generate code from a given flowchart image. We approach this task as a sequence-to-sequence generation problem involving two different modalities: image (flowchart) and text (code) and propose a framework, namely \model{} (\underline{F}lowchart-to-\underline{C}ode \underline{T5} Model) that involves: (i) reading and converting the flowchart image into a sequence encoding, and then (ii) autoregressively generating code using the flowchart encoding. Figure~\ref{fig:model} illustrates the proposed framework. We describe the two steps in the following subsections:

\subsection{Flowchart Encoding Generation}
In this step, we encode flowchart images into intermediate sequence encodings in the form of text. Given the flowchart image, we first detect and recognize the flowchart blocks, namely process (rectangle), decision (diamond), input/output (rhomboid), and terminal (oval), using the Hough transform-based shape detectors~\cite{opencv_library}. We further employ an off-the-shelf OCR method viz. easyOCR~\cite{easy_ocr} to recognize the text within the boxes and on arrowheads for digitized flowchart images. We then match the recognized shapes and text using their respective coordinates, i.e., a text is paired with the name of a block only if the text coordinates lie within the shape coordinates. The final flowchart encoding is a text sequence combining all the recognized text and shapes in the form of a key-value pair in the order in which they appear (from start to end). To this end, we experiment with three different strategies for encoding representation: (i) \emph{tuple encodings}, wherein each step of the flowchart is represented as a tuple of the text and the box shape, each within quotes; (ii) \emph{string encodings}, which eliminates quotes from text and shapes, and make use of braces to separate each step of the flowchart; and the optimized (iii) modified string encodings, where we utilize the [SEP] special tokens in the vocabulary of transformers to get rid of any additional braces or quotes and delineate each step of the flowchart. We provide an example for each of these encoding representations in Table~\ref{table:encodings}. We experiment with all three encoding forms and compare their effectiveness on our target task in Table~\ref{tab:encoding_results}.

\begin{table}[t!]
\caption{
Examples of different flowchart encoding. Refer to the main text for more details. A comparative study of different encodings is provided in Table~\ref{tab:encoding_results}.  
}
\centering
\resizebox{\linewidth}{!}{
 \def\arraystretch{1.25}%
\begin{tabular}{c | c | c}
\hline
\textbf{Tuple encodings} & \textbf{String encodings} & \textbf{Modified string encodings}
\\ \hline
&\\
\multirow{2}{*}{\parbox{6cm}{[('start fun1', 'OVAL'), ('input: X', 'PARALLELOGRAM'), ('y = ((16 + x) - 20)', 'RECTANGLE'), ('output: y', 'PARALLELOGRAM'), ('end function return', 'OVAL')]}}
&
\multirow{2}{*}{\parbox{6.5cm}{\{startfun1,OVAL\},\{input: X,PARALLELOGRAM\},\{y = ((16 + x) - 20),RECTANGLE\},\{output: y,PARALLELOGRAM\},\{end function return,OVAL\}}}
&
\multirow{2}{*}{\parbox{6cm}{start fun1, OVAL [SEP] input: X, PARALLELOGRAM [SEP] y = ((16 + x) - 20), RECTANGLE [SEP] output: y, PARALLELOGRAM [SEP] end function return, OVAL}}
\\ &\\& \\& \\ &\\
\hline
\end{tabular}
}
\label{table:encodings}
\end{table}

\begin{figure}[!t]
\caption{\textbf{Data augmentation:} Example of data augmentation used in code samples during pre-training of \model{}. We propose three logic-preserving augmentations that include changing the function names, variable names, and both. Augmented names are highlighted in red color.} \label{fig:data_augmentation}
\includegraphics[width=\textwidth]{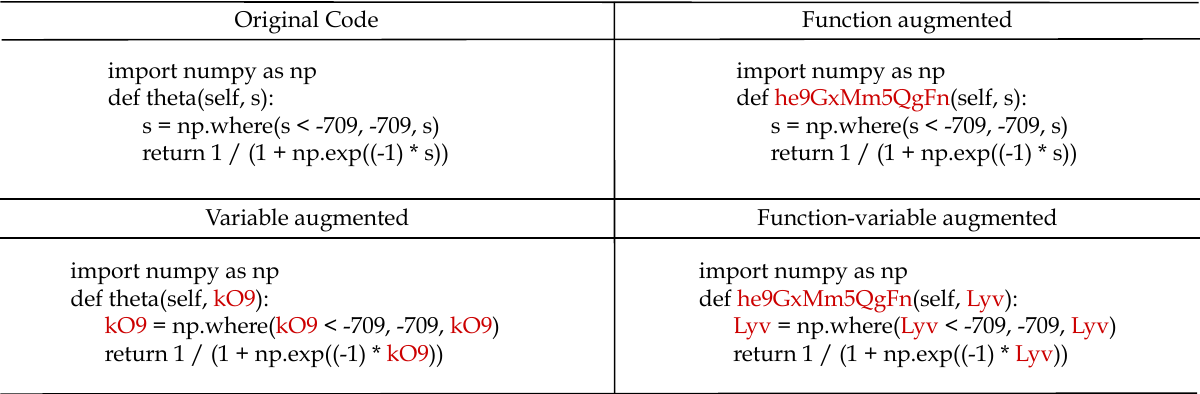}
\end{figure}

\subsection{Code Generation}
\begin{figure}[!t]
\caption{\textbf{Masked Modelling:} Example encoder inputs and decoder outputs during mask token prediction of \model{}.} \label{fig:pre_training}
\includegraphics[width=\textwidth]{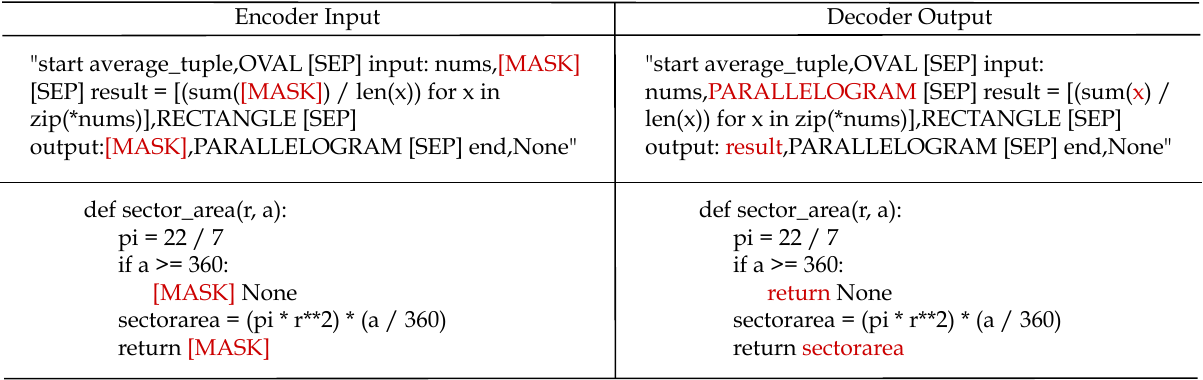}

\end{figure}

Inspired by the recent success of large-scale pre-trained code generation models, we adapt Code-T5~\cite{wang2021codet5} -- a transformer-based baseline trained for code generation, to our task. To this end, we initially pre-train it on a large number of logic-preserving augmented codes on the masked modeling objective in a self-supervised setting. The pre-training process adds knowledge of flowchart structure and code semantics to the model. Finally, we fine-tuned the pre-trained Code-T5 on the training set of \data{}. The data augmentation, pre-training, and fine-tuning process are performed as follows: \\  
\\
\textbf{Data Augmentation:} In order to increase the size of the dataset while keeping the logic of codes intact, we explored data augmentation. This has been achieved by changing function names and variable names. We augmented the training subset of the \data{} dataset. Replacing all functions and variables with a specific set of characters would make the dataset biased. Therefore, the function and variable names were constructed randomly using uppercase/lowercase letters, underscore, and/or digits while keeping the naming conventions for the Python programming language in mind. The length of the function names was chosen randomly from the range of $4-13$; for variable names, the range was $1-3$. Thus, each program was augmented in three different ways: changing the function or variable names or changing both function and variable names together. Figure~\ref{fig:data_augmentation} depicts all the augmentations corresponding to a sample code. After augmentation, the train dataset size increased from $10,102$ to $40,408$. These $30,306$ augmented codes have been utilized at the pre-training stage of our method.\\
\\
\textbf{Masked Modeling Objective:} Inspired by the success of the Masked Language Modeling (MLM) pre-training objective in BERT~\cite{kenton2019bert}, we propose an analogous objective specific to our problem. We adopted the pre-trained CodeT5 model and trained it on the augmented codes and flowchart encodings of the train set of \data{}. Tokens in the pre-training dataset are masked randomly at a probability of $0.15$, and we aim to optimize the loss associated with the reconstruction of the original sample, as shown below:
\begin{equation}
L_{mml}(E, \bar{E}) = -\sum_{t=1}^{N} log(e_t|e_{0:t-1}, \bar{E}).
\end{equation}
where  $E=<e_1,e_2\ldots,e_N>$ and $\bar{E}=<f_r(e_1), f_r(e2),\ldots,f_r(e_N)) >$  represent the ground truth and masked encodings/code, respectively. $\bar{E}$ is obtained by applying the function $f_r(e_i)$ to the ground truth encoding, which randomly replaces token $e_i$ with the mask token $[MASK]$ with a probability of $0.15$. $N$, $e_0$ denotes the length of the flowchart encoding and start token, respectively. Figure~\ref{fig:pre_training} shows examples of masked modeling implemented for encoding and a code sample. For the encoding input, if we mask the shape of a block (\emph{PARALLELOGRAM} in the given example), the model must be able to infer the correct shape based on the context and the pattern it has learned during training. 
 
\subsubsection{Fine-tuning:} 
After pre-training \model{} on augmented data, we further fine-tuned it on the training data of FloCo, for \task{} task. Figure~\ref{fig:model} (a) shows the training pipeline; the given flowchart image is first converted into sequence encoding by detecting shapes and the text inside the shapes using an off-the-shelf OCR technique. Positional encodings are added to the flowchart encodings before feeding them to the encoder. The decoder has access to the output of the encoder. It starts with a start token, and auto-regressively generates code token-by-token. To this end, during fine-tuning, we employed a language modeling loss expressed as follows:

\begin{equation}
L(X, E) = -\sum_{t=1}^{M}log(x_t|x_{0:t-1}, E).
\end{equation}
where $X = <x_1, x_2, \ldots, x_M>$ denotes the ground truth code. Additionally, $M$, and $x_0$ represent the length of the code and start token, respectively. Note that during both the pre-training and the fine-tuning stages, we include different flowchart box shapes as a special token in the transformer model's vocabulary.

\section{Experiments and Results}
In this section, we present an extensive experimental analysis on the \data{} benchmark to verify the efficacy of our proposed model.   
\label{experiment}
\subsection{Evaluation metrics}
Following the code generation literature~\cite{ahmad2021unified}, we evaluated the performance of our baselines and proposed model using the following three metrics:
\begin{enumerate}[i.]
    \item \textbf{BLEU~\cite{papineni2002bleu}}: a widely used word-overlap metric for assessing the quality of machine-translated text by comparing the $n$-grams of the generated code to the reference (ground truth) code and counting the number of matches. 
    \item \textbf{CodeBLEU~\cite{ren2020codebleu}}: a specialized metric that evaluates the quality of generated code, taking into account syntactical and logical correctness and the code's structure as reflected in the abstract syntax tree and data flow, in addition to comparing n-grams.
    \item \textbf{Exact Match (EM)}: a binary metric that checks if the generated code sequence is exactly the same as the ground-truth code.
\end{enumerate}

\subsection{Baseline Models}
To evaluate the effectiveness of the proposed method, we compared it against the following four competitive baselines:\\
\noindent \textbf{Vanilla Transformer~\cite{vaswani2017attention}} is the attention-based encoder-decoder architecture upon which the transformer-based pre-trained models are built. By comparing the proposed method with this baseline, we can observe the specific advantages of pre-training.\\
\noindent \textbf{BART~\cite{lewis2020bart}} is a pre-trained, bidirectional, autoregressive encoder-decoder architecture that was pre-trained on unlabelled natural language data and optimized using reconstruction loss. The noising techniques used were token masking, token deletion, text infilling, sentence permutation, and document rotation.\\
\noindent \textbf{PLBART~\cite{ahmad2021unified}} is an extension of BART and was pre-trained on a large-scale dataset containing unlabelled natural language and programming language data. The pre-training objective was denoising autoencoding, and the noising strategies used were token masking, deletion, and infilling.\\
\noindent \textbf{CodeT5~\cite{wang2021codet5}} adopted the T5 (pre-trained on natural language) architecture and was pre-trained on natural language and programming language data. The pre-training objectives were span prediction, identifier tagging, masked identifier prediction, and bimodal dual generation.

By comparing our proposed method with these baselines, we can observe how our method outperforms them and understand how it leverages the pre-training.

\begin{table}[!t]
\centering
\caption{On the \data{} test set, we compared \model{} to competitive transformer-based baselines and found that our method achieved higher scores for all metrics.}\label{tab:main_res}
\begin{tabular}{l c c c}
\toprule
Method~~~~&~~~~BLEU~~~~&~~~~CodeBLEU~~~~&~~~~EM~~~~\\
\midrule
Vanilla Transformer~\cite{vaswani2017attention} & 10.3 & 26.8 & 0.0 \\
BART~\cite{lewis2020bart} & 31.1 & 40.7 & 2.2\\
PLBART~\cite{ahmad2021unified} & 55.7 & 63.7 & 19.4 \\
CodeT5~\cite{wang2021codet5} & 63.8 & 71.8 & 17.8 \\
\midrule
\textbf{\model{}} & \textbf{67.4} & \textbf{75.7} & \textbf{20.0}\\
\bottomrule
\end{tabular}
\end{table}

\begin{figure}[!tp]
\includegraphics[width=\textwidth]{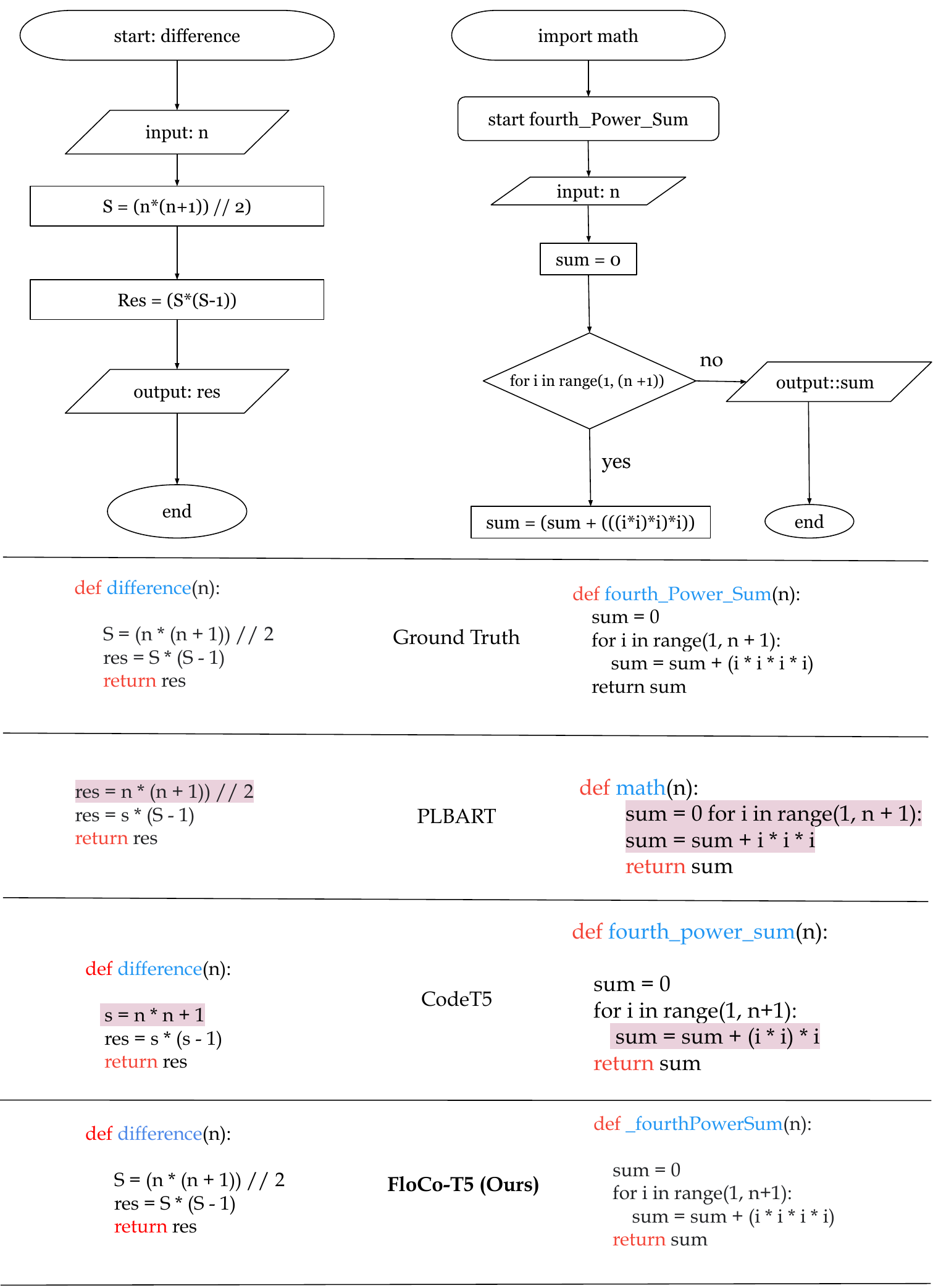}
\caption{\textbf{Qualitative comparison:} Results on two flowchart images using PLBART, CodeT5 and our method. Errors are highlighted in dark red color. The codes generated by our method are similar to the ground truth as compared to the PLBART and CodeT5. CodeT5 has fewer errors as compared to PLBART.} \label{fig:qaul_res}
\end{figure}

\subsection{Implementation details for Reproducibility}
\model{} is implemented using the Huggingface library~\cite{wolf-etal-2020-transformers} and utilizes the implementation of CodeT5~\cite{wang2021codet5}, using the `Salesforce/codet5-base' pre-trained model available on Huggingface. The model contains 222.9 million trainable parameters. It consists of $12$ encoder, $12$ decoder layers, and $12$ attention heads in each layer. The input encodings are truncated or padded to a maximum length of 512 tokens. We optimize training using the Adam ~\cite{kingma2014adam} optimizer with a learning rate of $1e-5$, a warmup for $2450$ steps, and a batch size of $16$. We use the same training configuration in both the pre-training and fine-tuning stages. All the baselines were trained on a single NVIDIA A6000 GPU with $48$ GB VRAM. The training process requires nearly 12 hours to reach convergence. We make our implementation available here: \url{https://vl2g.github.io/projects/floco/}. 

\begin{table}[!t]
\centering
\caption{Comparision of different flowchart encoding representations on the performance of \model{}.}\label{tab:encoding_results}
\begin{tabular}{l c c c}
\toprule
Method~~~~&~~~~BLEU~~~~&~~~~CodeBLEU~~~~&~~~~EM~~~~\\
\midrule
Tuple encodings & 16.7 & 37.7 & 0.2\\
String encodings & 50.1 & 63.4 & 11.1\\
\textbf{Modified string encodings} (Ours) & \textbf{67.4} & \textbf{75.7} & \textbf{20.0} \\
\bottomrule
\end{tabular}
\end{table}

\subsection{Results and discussions}
We evaluate our method on the proposed \data{} dataset and compare it against competitive baselines, namely vanilla transformer~\cite{vaswani2017attention}, BART~\cite{lewis2020bart}, PLBART~\cite{ahmad2021unified}, and CodeT5~\cite{wang2021codet5}.  Table~\ref{tab:main_res} shows the performance of the implemented baselines and proposed \model{} on three evaluation metrics. Vanilla Transformer~\cite{vaswani2017attention} is trained from scratch in contrast to other baselines, pre-trained on large-scale unlabelled data with different self-supervised pre-training objectives. Hence, Vanilla Transformer lacks the understanding of language and programming semantics and structure, resulting in the lowest performance for all the metrics. BART~\cite{lewis2020bart} is pre-trained on natural language text and thus, has a better understanding of the semantics and structure of the sequential data, as natural text also has rules, structure, and other syntactical properties. It results in better performance as compared to the Vanilla Transformer for all of the metrics. PLBART is pre-trained on the natural text and programming language, which means it has a better understanding of code structure and semantics, resulting in better performance compared to BART and Vanilla Transformer on all metrics. CodeT5~\cite{wang2021codet5} is pre-trained with programming-language-specific, fine-grained identifier-aware denoising tasks, which help in exploiting code semantics and structure in a more exquisite way, resulting in significant improvement over other baselines. In the proposed \model{}, we adopted a pre-trained CodeT5 model, which has task-specific knowledge, and further pre-trained it on augmented training samples for the mask token generation task. As expected, \model{} outperforms all baselines for all the metrics used for evaluation, showing the efficacy of the proposed code augmentation and pre-training strategy.

Figure~\ref{fig:qaul_res} shows the generated codes for two flowchart samples. We compare the ground truth codes with the ones generated from PLBART, CodeT5, and our method. \model{} is able to generate codes syntactically correct codes, which are similar to the ground truth codes, while other baselines fall short in generating correct codes. This observation is same across other test samples as numerically summarized by Table~\ref{tab:main_res}.

We further conducted an experiment with three flowchart image encoding methods, shown in Table~\ref{table:encodings}, and results presented in Table~\ref{tab:encoding_results}. The modified string encoding method utilized a [SEP] token to separate each step of the flowchart, and removed extra braces, enhancing the preservation of the flowchart's structure, and consequently outperforming other encoding methods.

\begin{figure}[!t]
\centering
\includegraphics[width=\textwidth]{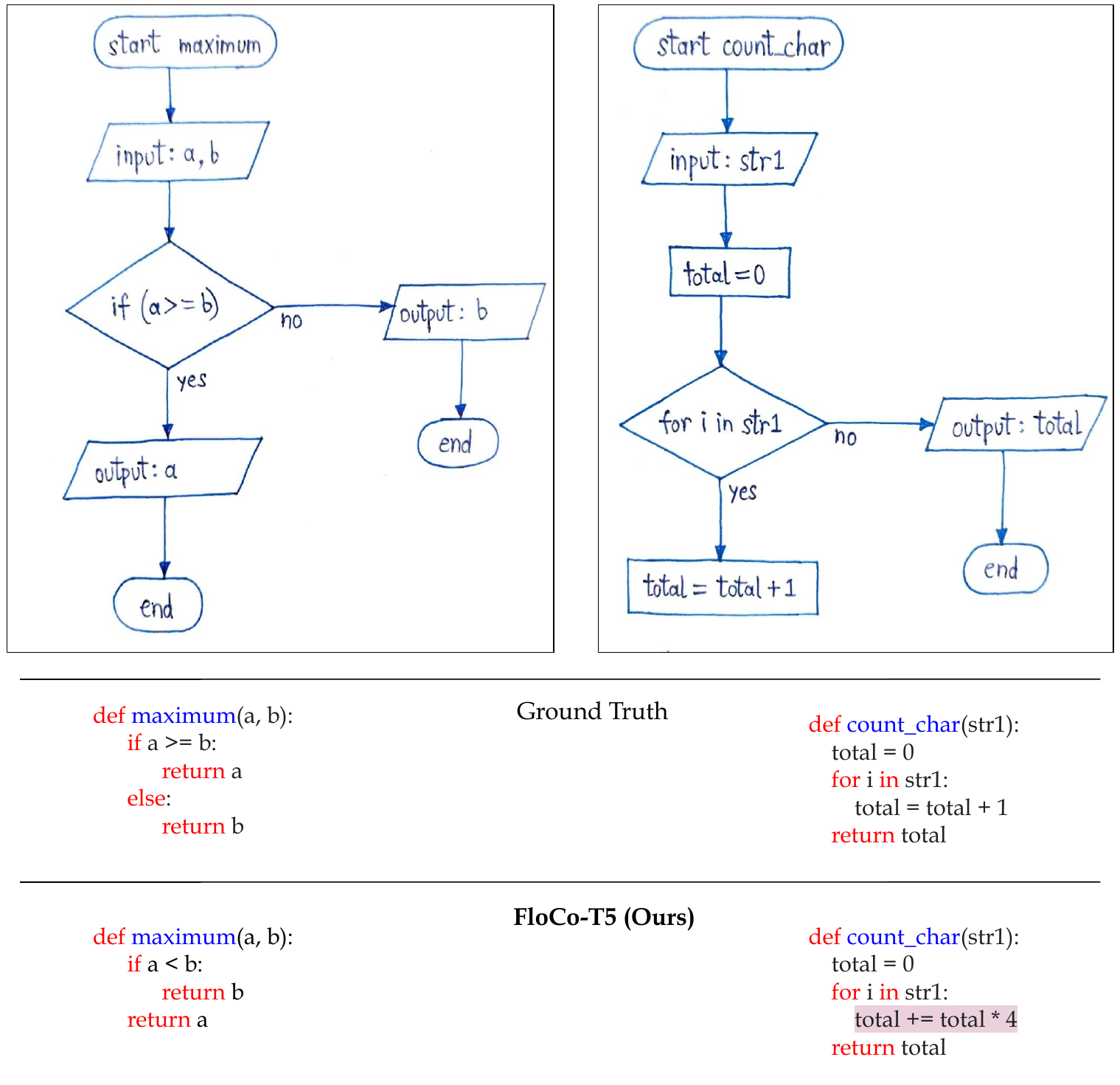}
\caption{\textbf{} Results on hand-drawn flowchart images using FloCo-T5. Errors are highlighted in dark red color. We observe an error in the program due to incorrect recognition of the handwritten character `*' by our OCR module.} \label{fig:handwritten_res}
\end{figure}

\begin{figure}[t]
\centering
\includegraphics[width=0.9\textwidth]
{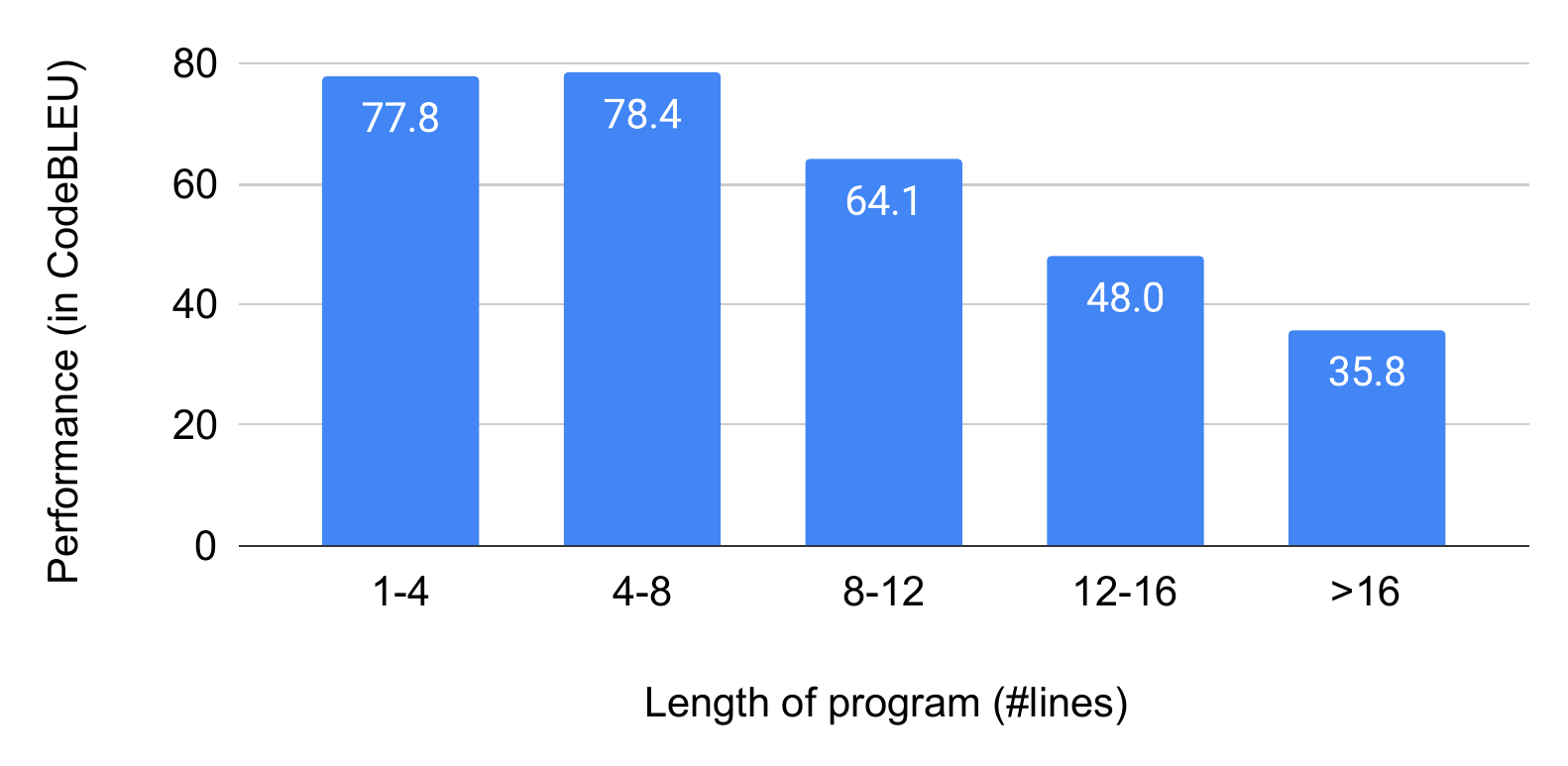}
\caption{Performance of \model{} across various program lengths.} \label{fig:performance_vs_program_length} 
\end{figure}

\noindent\textbf{Can the proposed approach work for hand-drawn Flowchart Images?}
We evaluated \model{} on hand-drawn flowchart images using $40$ samples created by three human annotators. Flowchart block detection and recognition were performed with OpenCV~\cite{opencv_library}. For handwritten text recognition, we employed CRAFT text detection~\cite{baek2019CRAFT} and TrOCR text recognition~\cite{li2021trocr}. \model{} achieved a BLEU score of 21.4\% and a CodeBLEU score of 34.6\% on these hand-drawn flowcharts. Fig.~\ref{fig:handwritten_res} displays Python codes generated for two sample hand-drawn flowcharts. These results indicate our approach's suitability for hand-drawn flowcharts, and performance can be significantly enhanced with advances in handwritten text recognition.
\newline
\noindent\textbf{Limitations:} We observed that code generation performance of our model is higher for shorter programs ($< \approx 12$ lines) but drops for longer programs ($> \approx 15$ lines) due to the dataset's bias towards shorter programs (average length: 4.6 lines) as shown in Figure~\ref{fig:performance_vs_program_length}. To address this issue, we propose increasing the number of training samples for longer programs. Future work will focus on expanding the \data{} dataset to include longer and more complex code and researching information flow in state and block diagrams.

\section{Conclusion}
We introduced the \model{} framework for generating Python code from flowchart images and presented the \data{} dataset to benchmark the \task{} task. \task{} is modeled as a sequence-to-sequence problem, where flowcharts are first encoded by detecting the shapes of blocks and reading the text within, and further transformed into code using competitive transformer baselines. \model{}'s task-specific pre-training results in significant improvements over related baselines. The recent advancements in Large Language Models (LLMs), such as ChatGPT, have revolutionized the field of code generation, and they can be adapted to solve our task. However, ensuring that these massive models have not seen our test data is not a trivial task. Furthermore, despite these advancements, we firmly believe that our dataset can be used to study open problems such as development of lightweight and interpretable models for generating code from flowchart images. We leave these as future directions to work on.

\noindent\textbf{Acknowledgements:} This work was partly supported by MeitY, Govt. of India (project number: S/MeitY/AM/20210114). Yogesh Kumar is supported by a UGC fellowship. 

{\small
\bibliographystyle{splncs04}
\bibliography{ref}
}

\end{document}